\documentclass{article}

\usepackage[final]{corl_2020} 

\usepackage[utf8]{inputenc} 
\usepackage[T1]{fontenc}    
\usepackage{url}            
\usepackage{booktabs}       
\usepackage{amsfonts}       
\usepackage{nicefrac}       
\usepackage{microtype}      

\usepackage{siunitx}
\usepackage{mathtools}
\usepackage{booktabs}
\usepackage{multicol}
\usepackage{multirow}
\usepackage{subcaption}
\usepackage{graphicx}
\usepackage{xcolor}
\usepackage{pifont}
\usepackage{tabularx}
\usepackage{bbm}

\DeclareFontFamily{OMX}{yhex}{}
\DeclareFontShape{OMX}{yhex}{m}{n}{<->yhcmex10}{}
\DeclareSymbolFont{yhlargesymbols}{OMX}{yhex}{m}{n}
\DeclareMathAccent{\wideparen}{\mathord}{yhlargesymbols}{"F3}

\newcommand{\ourmodel}{Streaming Object Detection \\from LiDAR Packets} %
\newcommand{\ourmodelshort}{\textsc{StrObe}}
\newcommand{\ourdataset}{\textsc{PacketATG4D}}

\usepackage{floatrow}
\DeclareFloatVCode{beforefloat}{\vspace{4pt}}
\DeclareFloatVCode{afterfloat}{\vspace{-8pt}}
\DeclareFloatVCode{beforefloatfig}{\vspace{4pt}}
\DeclareFloatVCode{afterfloatfig}{\vspace{-12pt}}
\newcommand{\cutsectionup}{\vspace*{-6pt}}
\newcommand{\cutsectiondown}{\vspace*{-2pt}}

\newcommand{\cutsubsectionup}{\vspace*{-6pt}}
\newcommand{\cutsubsectiondown}{\vspace*{-2pt}}

\newcolumntype{s}{>{\centering\arraybackslash}X}

\graphicspath{{imgs/}}

\title{\ourmodelshort{}: \ourmodel{}}

\author{%
    \textbf{Davi Frossard} \textsuperscript{1,2} \quad
    \textbf{Simon Suo} \textsuperscript{1,2} \quad
    \textbf{Sergio Casas} \textsuperscript{1,2} \quad
    \textbf{James Tu} \textsuperscript{1,2} \quad\\
    \textbf{Rui Hu} \textsuperscript{1} \quad
    \textbf{Raquel Urtasun} \textsuperscript{1,2}\\
    \textsuperscript{1} Uber Advanced Technologies Group \quad
    \textsuperscript{2} University of Toronto\\
    \texttt{\{frossard, suo, sergio.casas, james.tu, rui.hu, urtasun\}@uber.com}
}

\begin{document}

\maketitle

\begin{abstract}
    Many modern robotics systems employ LiDAR as their main sensing modality due to its geometrical richness. Rolling shutter LiDARs are particularly common, in which an array of lasers scans the scene from a rotating base. Points are emitted as a stream of packets, each covering a sector of the \ang{360} coverage. Modern perception algorithms wait for the full sweep to be built before processing the data, which introduces an additional latency. For typical 10Hz LiDARs this will be 100ms. As a consequence, by the time an output is produced, it no longer accurately reflects the state of the world. This poses a challenge, as robotics applications require minimal reaction times, such that maneuvers can be quickly planned in the event of a safety-critical situation. In this paper we propose \ourmodelshort{}, a novel approach that minimizes latency by ingesting LiDAR packets and emitting a stream of detections without waiting for the full sweep to be built. \ourmodelshort{} reuses computations from previous packets and iteratively updates a latent spatial representation of the scene, which acts as a memory, as new evidence comes in, resulting in accurate low-latency perception. We demonstrate the effectiveness of our approach on a large scale real-world dataset, showing that \ourmodelshort{} far outperforms the state-of-the-art when latency is taken into account, and matches the performance in the traditional setting.
\end{abstract}
\cutsectionup
\section{Introduction}
\label{sec:introduction}
\cutsectiondown

\begin{figure}[t]
    \centering
    \includegraphics[width=0.95\linewidth]{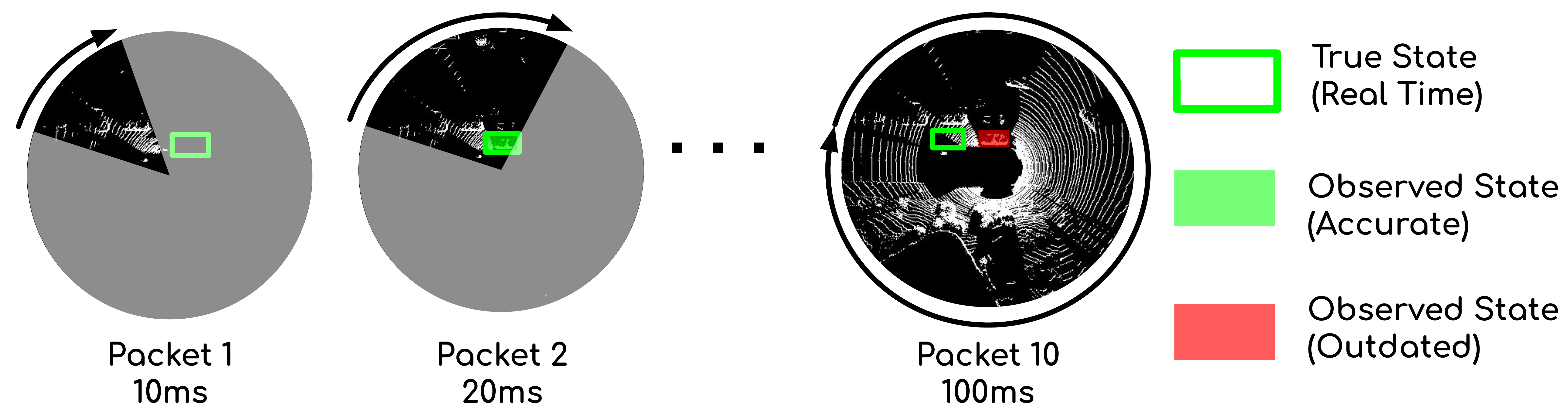}
    \caption{Objects are observed at different times when building a full LiDAR sweep, indicated as the solid boxes. If a full sweep from a 10hz LiDAR is accumulated before detection, a latency of 100ms will be introduced and by the time a detection is available (Packet 10) it no longer reflects the state of the world.}
    \label{fig:latency_teaser}
\end{figure}

Perceiving the world is a critical task in modern robotics applications. Self-driving vehicles must first process sensory information to perform object detection and estimate the free space  before attempting to plan a safe and comfortable maneuver towards the goal. LiDAR has become the main sensing modality in most self-driving vehicles due to the geometrical richness it provides. Most prevalent LiDAR sensors operate by collecting a rotating scan of the environment, typically completing revolutions at a 10hz rate. However, as the sensor rotates, observations arrive as a stream of spatio-temporal points $(x, y, z, t)$ grouped in fine-grained \emph{packets}, each spanning approximately 10ms. This gives rise to a rolling shutter effect shown in \autoref{fig:latency_teaser}, where objects in different locations are observed asynchronously.

Modern autonomous systems accumulate the LiDAR packets into a full \ang{360} sweep before running perception. This waiting time adds significant latency to the  pipeline, particularly for objects that were seen in the earlier packets in the sweep. It also introduces an erroneous assumption that all observations in the full sweep are made synchronously. In reality, when the perception model receives the input, there is already a discrepancy between the outdated observations and the true state of the world, illustrated in \autoref{fig:latency_teaser}. Furthermore, there is a  temporal discontinuity in the sweep where the earliest and the latest packets meet which creates artifacts in the point cloud.

For safety-critical applications like self-driving, even minimal delays  may result in catastrophic outcomes. For example, in the presence of high-speed vehicles, building a sweep from a 10Hz LiDAR  introduces a latency of 100ms, which translates to several meters of error in free space estimation.  Having lower latency is crucial in safety-critical situations where the vehicle must quickly perceive and react to avoid harmful events. Therefore, it is important to process incoming sensory information with minimal latency.

Processing individual LiDAR packets can be challenging, since only a small sector of the scene is observable as illustrated in \autoref{fig:latency_teaser}. 
Objects of interest are often fragmented across different LiDAR packets, particularly when close to the sensor. Coincidentally, that is also when high accuracy and low latency are the most important as \emph{close range} objects are typically the most critical to safety. Thus, individual packets alone may be insufficient for high quality detections, making it necessary to incorporate past observations. 

Existing LiDAR object detectors generally assume access to a full 360 degree sweep, or a large subregion (e.g., front view) that spans all objects of interest. As such, these models do not explicitly reason about objects split across multiple observations. As shown in our experiments, directly adopting full-sweep models for processing individual LiDAR packets is not a good solution due to the partial observation and lack of global context. Conversely, exploiting multiple sweeps ~\cite{casas2018intentnet,luo2018fast} provides richer geometrical evidence as more LiDAR points are collected over time. However, most current solutions are computationally inefficient as each packet would be processed as many times as the duration of the history. As such, naively aggregating historical sensory information at the input level is not amenable to emitting low latency object detections from fine-grained LiDAR packets. 

In this paper we propose \ourmodelshort, a novel detection model which exploits the sequential nature of LiDAR observations and efficiently reuses past computation to stream low latency object detections from LiDAR packets. Our approach voxelizes the incoming LiDAR packets into a Bird’s-Eye View (BEV) grid, and uses an efficient convolutional backbone to process only the relevant region. Furthermore, we introduce a multi-scale spatial memory that is read and updated with each LiDAR packet. This allows us to reuse past computation, and make the incremental processing of incoming LiDAR packets lightweight and efficient. Importantly, we achieve an end-to-end latency of 21 ms (from observing an actor to emitting a detection) on an NVIDIA 2080Ti: 10ms for accumulating a packet and 11ms for model inference. In contrast, even fast full sweep detectors \cite{yang2018pixor} operate at an order of magnitude higher latencies: Taking 100ms to accumulate the sweep and another 28ms for model inference, for a total of 128ms.

Our second contribution is a novel large scale benchmark for evaluating streaming object detection from LiDAR packets. Unlike existing public datasets, \ourdataset{} contains LiDAR data at the packet level, along with accurate ego-pose and associated object bounding box annotations at the same temporal resolution (i.e., 100Hz). We also propose a novel metric \emph{latency-aware mAP} to explicitly take latency into account when evaluating perception. We show that our approach far outperforms the state-of-the-art when the data buffering latency is taken into account, while still matching the performance in the conventional setting. 

\cutsectionup
\section{Related Work}
\label{sec:related_work}
\cutsectiondown

3D object detection has made tremendous progress in recent years due to the advances of deep learning and the availability of large-scale labeled datasets. The topic  of how to effectively process LiDAR data has received significant attention and many approaches have been proposed. Point clouds have been processed in perspective format using a range image~\cite{li2016fcn,meyer2019lasernet}. By converting the point cloud into an image, these approaches can leverage the vast body of knowledge on 2D object detection to build good architectures for the task. However, such methods suffer from the same challenge present in 2D detection: high variance in receptive field requirements as a function of depth.

To tackle these issues, some methods perform 3D detection directly on the unstructured 3D points. This is usually achieved through first extracting local signatures with a fully connected layer~\cite{qi2017pointnet,li2018pointcnn,hua2018pointwise,wang2018deep} or by using deformable filters ~\cite{xiong2019deformable}. An alternative framework is to voxelize the points into a regularly spaced 3D grid, making reasoning on point clouds amenable to convolutional architectures. Early works ~\cite{wu20153d,maturana2015voxnet,li20173d,luo2018fast} leverage 3D convolutions, but they are memory intensive. Others \cite{riegler2017octnet,ren2018sbnet,yan2018second} exploit the sparsity of point clouds to reduce redundant computation and make higher resolution processing feasible. BEV detectors \cite{yang2018pixor,yang2018hdnet,casas2018intentnet} avoid heavy computation by exploiting efficient 2D convolutions over a top-down pseudo-image of the scene. Other methods have leveraged hybrid representation of points and voxels \cite{chen2019fast,shi2020pv,yang2019std,lang2019pointpillars,zhou2020end} to exploit the benefits of both representations. 

However, the aforementioned methods assume a full sweep is available, which requires the sensor to complete a full rotation and incurs latency. Previous works have explored the problem of latency in different settings, for instance on the effect of model runtime for 2D object detection~\cite{li2020towards}, or how the temporal aspect of point clouds is relevant for odometry and mapping~\cite{alismail2014continuous, zhang2017low}. Concurrent work \cite{han2020streaming} has considered streaming object detections from a rolling shutter LiDAR. However, their model uses an LSTM to maintain the state, which does not leverage the spatial nature of the problem. Furthermore, their evaluation does not capture the impact latency has on the accuracy of state estimation.
\cutsectionup
\section{Low Latency Detection on Streaming LiDAR}
\label{sec:methodology}
\cutsectiondown

\begin{figure}[tb] 
    \centering
    \includegraphics[width=\textwidth]{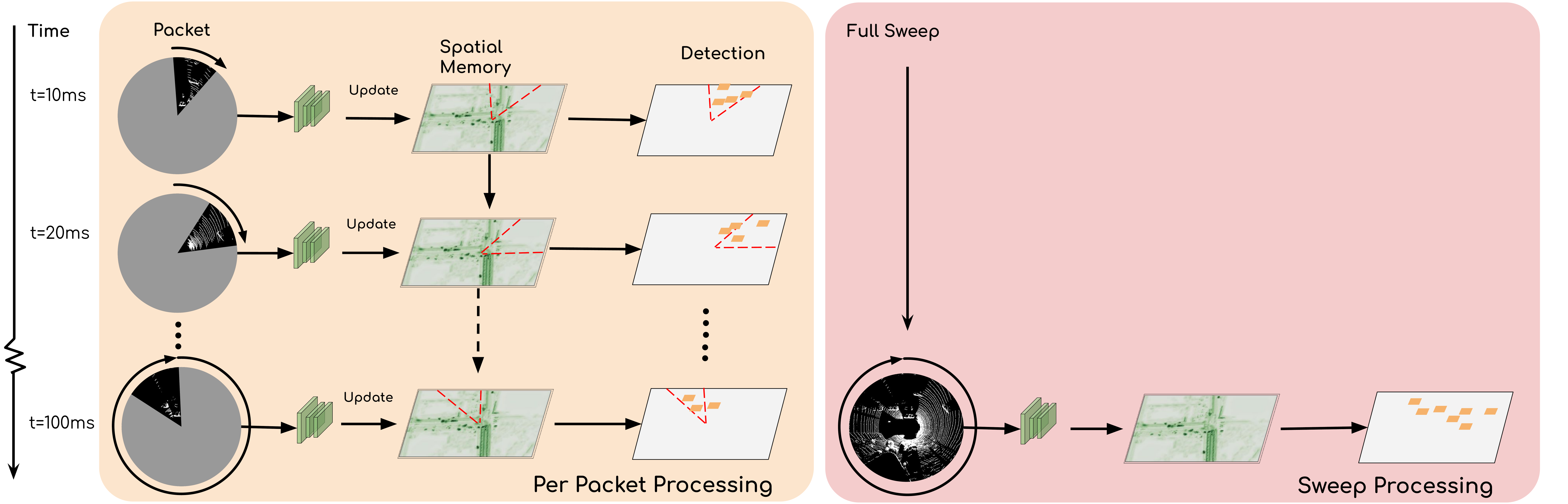}
    \caption{Existing point cloud perception methods wait 100ms to accumulate the full sweep (\textbf{right}). \ourmodelshort{} (\textbf{left}) is able to process each packet and emit new detections with high accuracy and minimal latency, while leveraging global context by continuously updating a spatial memory that keeps track of previously seen packets.}
    \label{fig:runtime_diagram}
\end{figure}

In this paper, we propose \ourmodelshort{}, a low-latency object detector that emits detections from streaming LiDAR observations. As illustrated in 
\autoref{fig:runtime_diagram}, as the LiDAR sensor spins, it yields data in sector packets (each roughly spanning \ang{36} in our 10Hz LiDAR). As opposed to previous models, which buffer this data into a full sweep before processing, our proposed method operates at the packet level. In doing so, we lower our latency by 90ms. A fundamental component to our approach is a novel spatial memory module design to reuse past computation, and make incremental processing of incoming LiDAR packets lightweight and effective. 

\subsection{Streaming Object Detection}

\begin{figure}[tb] 
    \centering
    \includegraphics[width=\textwidth]{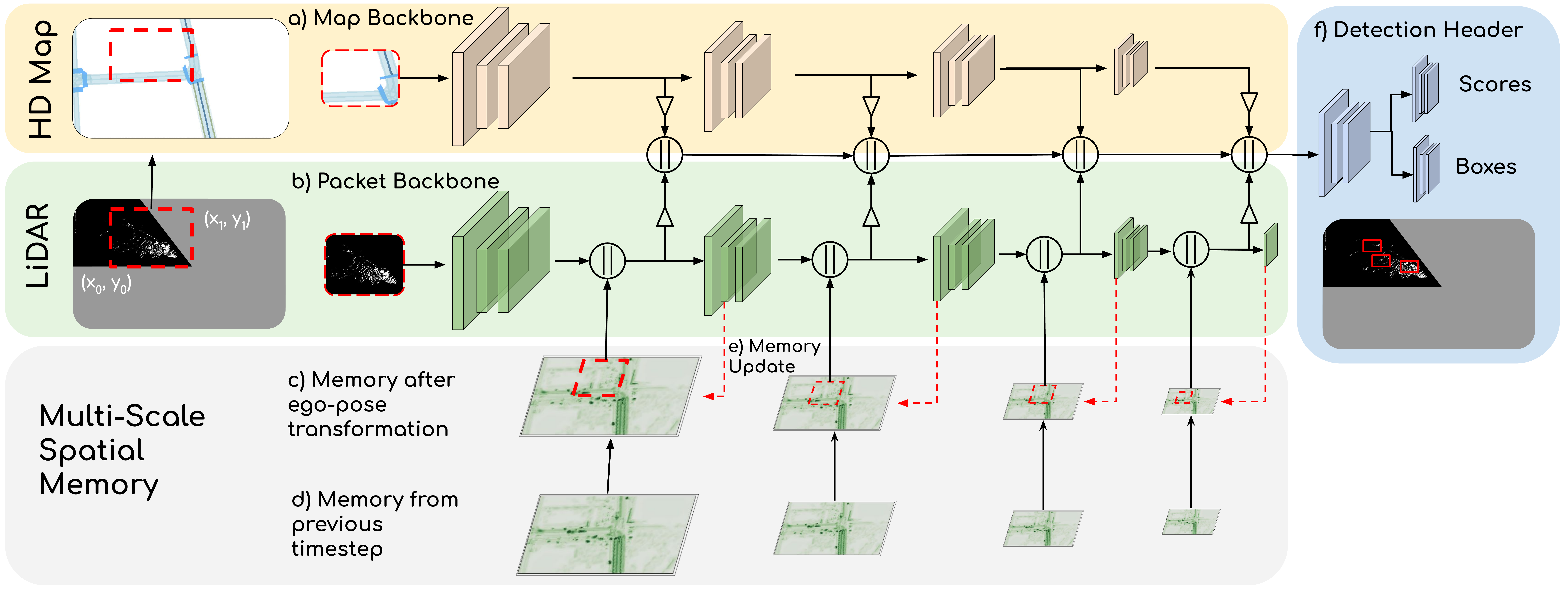}
    \caption{\textsc{\ourmodelshort{}} performs regional convolution on LiDAR packets and HD maps, using a multi-scale spatial memory for global reasoning. $\bigtriangleup$ is interpolation and $\parallel$ channel-wise concatenation.}
    \label{fig:network}
\end{figure}

The overall architecture of our model is illustrated in \autoref{fig:network}. The network takes as input a LiDAR packet and an HD map, which is useful as a prior on the location of actors (e.g., a vehicle is more likely to be on the road than on the sidewalk). For each packet we first voxelize the points and rasterize into a BEV pseudo-image with height as the channel dimension \cite{yang2018pixor}. Following \cite{yang2018hdnet,casas2018intentnet}, we also rasterize the map into a BEV pseudo-image, where each channel corresponds to a different layer of the map (e.g., crosswalks, roads, etc).  We then extract features using our novel   regional convolutions (\autoref{fig:network} -- a, b), which only compute features in the rectangular area defined by the packet. A latent spatial representation of the scene is then maintained using a memory module (\autoref{fig:network} -- c, d, e). Lastly, we channel-wise concatenate multi-scale features and regress detection parameters using our output header (\autoref{fig:network} -- f).

\paragraph{Regional Convolution Layer:} To reduce latency while leveraging the proven strength of BEV representations and 2D convolutions, we propose to process the input with a local operator, which we call regional convolution. Specifically, for an input $x$ and coordinates $x_0, x_1, y_0$ and $y_1$, we extract features $\mathbf{y}$ only on the region $\mathbf{x}[x_0:x_1, y_0:y_1]$, where the brackets denote indexing at the rectangle defined by the coordinate ranges. This allows us to leverage locality to minimize wasted computation.
\begin{gather}
    \mathbf{y} = f_\text{region}\left(\mathbf{x}[x_0:x_1, y_0:y_1], \mathbf{w}\right)
    \label{eq:region_conv}
\end{gather}
In practice, $f_\text{region}$ is a sequence of 2D convolution, ReLU activation and Group Normalization \cite{wu2018group}. Furthermore, for both the LiDAR packet and HD map, the region coordinates are defined as the minimal rectangle that fully encloses all points in the LiDAR packet. This is illustrated in \autoref{fig:network} -- a, b. 

\paragraph{Spatial Memory:} While regional convolutions allow us to efficiently ingest packets, independently processing them is not sufficient for accurate perception since objects will often be fragmented across many packets. Furthermore, a single observation of an object far away will typically yield few points due to the sparsity of the sensor at range. We would thus like to  leverage information from previous scans of the region. However, naively processing the history of observations every time we receive a packet results in redundant computation and slow inference. Instead, our approach iteratively builds a global spatial memory from a series of partial observations while at the same time producing new detections with each LiDAR packet, \autoref{fig:network} -- c. This enables us to re-use past computation and produce low-latency and accurate detections. Importantly, the LiDAR points are registered on a consistent coordinate frame defined by a continuous ego-pose. The memory is aligned with this pose by bilinearly resampling its features to account for ego-motion with every new packet (\autoref{fig:network} -- c, d). This guarantees that the LiDAR and map features are consistently aligned with the spatial memory in the same coordinate frame.

\paragraph{Memory Update:} As each LiDAR packet arrives, the spatial memory is incrementally updated with new local features to reflect the latest state (\autoref{fig:network} -- d). Each update step is done through aggregation of the current memory state $\mathbf{m}$ and the incoming local features $\mathbf{y}$. Specifically, we employ a channel reduction with learned parameters $\mathbf{w}$ as follows 
\begin{gather}
    \label{eq:memory_update}
    \mathbf{m'}[x_0:x_1, y_0:y_1] = f_\text{memory}\left(\mathbf{m}[x_0:x_1, y_0:y_1], \mathbf{y}, \mathbf{w}\right).
\end{gather}
In practice, $f_\text{memory}$ channel-wise concatenates $\mathbf{m}$ and $\mathbf{y}$, resulting in a tensor with $2c$ channels, then applies two blocks of 2D convolution, ReLU activation and Group Normalization, with the second block bringing the number of channels back to $c$. This is illustrated as the red dotted arrows in \autoref{fig:network} -- e.

\paragraph{Multi-Scale Backbone:} In order to leverage the semantic representations of feature maps at different scales (i.e., richer geometry on higher resolutions; richer semantics on lower) we employ a multi-scale backbone for the extraction of both LiDAR and HD map features. Together with the spatial memory at each scale, the benefits of this are twofold: It allows the model to regress accurate and low latency detections from partial observations by remembering the features from immediately preceding packets. It also makes it possible for the network to persist long term features that are useful to detect objects through occlusion over multiple sweeps as well as overwrite previous features when stronger evidence is available.

\paragraph{Architecture Details:} We employ a  BEV grid  with resolution of 0.2m for each voxel. This grid then   goes through 4 blocks of [2, 2, 3, 6] Regional Convolution layers with [24, 64, 128, 256] channels, followed by Max Pooling with a stride of 2. Each block has a corresponding Spatial Memory that holds the pre-pooling state of the features. In parallel, features are extracted from the HD map with a backbone that consists of a sequence of 4 blocks with [2, 2, 3, 3] Regional Convolution layers with [16, 32, 64, 128] channels. After each block, Max Pooling with a stride of 2 is employed. The feature maps from each block of both the LiDAR and HD map backbones are then bilinearly resized to a common resolution of 0.8m, channel-wise concatenated, and processed by one last block of 4 Regional Convolutions with 256 channels. 

\paragraph{Detection Header:} We perform multi-class BEV detection for vehicles, cyclists, and pedestrians via a single-stage detection header consisting of 2 convolutional layers that predict the classification and regression targets for each cell in the fused feature map (hereinafter referred to as "anchors"). 
All objects are defined via their centroid $(b_x, b_y)$ and confidence $\sigma$, whereas cyclists and vehicles also have length, width, and heading $(b_l, b_w, b_\phi)$ in BEV. For the confidence, we predict its logit $\log \frac{\sigma}{1-\sigma}$. We define the centroid of the box $(b_x, b_y)$ as an offset $(\Delta x, \Delta y)$ from the coordinates of the center point of its anchor pixel $(a_x, a_y)$:
\begin{gather}
    (b_x, b_y) = (a_x + \Delta x, a_y + \Delta y).
\end{gather}

For the vehicle dimensions we  predict $[\log l, \log w]$, which encourages the network to learn a prior on the dimension of the boxes (low variance should be expected from the dimension of vehicles). The heading $b_\phi$ is parameterized by the tangent value. In particular, we predict a signed ratio so that the specific quadrant can be retrieved:
\begin{gather}
    b_\phi = \arctan\frac{\theta_1}{\theta_2}.
\end{gather}

\cutsubsectionup
\subsection{Learning}
\cutsubsectiondown

Following common practice in object detection \cite{ren2015faster}, we employ a multi-task loss over classification and bounding box regression to optimize the model (using $\alpha = 2.0$), i.e:
\begin{gather}
    \mathcal{L} = \mathcal{L}_\text{reg} + \alpha \mathcal{L}_\text{cls}.
\end{gather}

\paragraph{Regression Loss:}
It is defined as the weighted sum of the smooth $\ell_1$ loss \cite{DBLP:journals/corr/Girshick15} between the ground truth vector of detection parameters  $\hat{\mathbf{y}} = [\Delta x, \Delta y, \log w, \log l, \theta_1, \theta_2]$ and predictions $\mathbf{y}$ with  $\gamma = [1, 1, 1, 1, 2, 2]$. Note that $\log w, \log l, \theta_1$ and $\theta_2$ are not considered for pedestrians since we are only concerned with predicting their centroid. 
\begin{gather}
    \mathcal{L}_\text{reg}(\mathbf{y}, \hat{\mathbf{y}}) = \dfrac{1}{N} \sum_{i=0}^N \gamma \cdot \mathrm{smooth}_{\ell 1}(\mathbf{y}^i_d - \hat{\mathbf{y}}^i_d)
\end{gather}

\paragraph{Classification Loss:}
It is defined as the binary cross entropy between the predicted scores and the ground truth. Due to severe class imbalance between positive $\hat{\mathbf{y}}_\text{pos}$ and negative $\hat{\mathbf{y}}_\text{neg}$ anchors given that most pixels in the BEV scene do not contain an object, we employ hard negative mining:
\begin{gather}
    \mathcal{L}_\text{cls}(\mathbf{y}, \hat{\mathbf{y}}) = \dfrac{1}{N} \sum_{i=0}^N \hat{\mathbf{y}}^i_\text{pos} \log \mathbf{y} + \dfrac{1}{K} \sum_{i=0}^N \mathbbm{1}[i \in \mathcal{N}_K](1 - \hat{\mathbf{y}}_\text{neg}^i) \log(1 - \mathbf{y})
\end{gather}
where $\mathcal{N}_K$ is a set containing $K$ hard negative anchors. This is obtained by sampling 750 anchors for vehicles, 1500 for cyclists and pedestrians, and picking the 20 with highest loss for each class.

\paragraph{Sequential Training:}
Due to the sequential nature of the memory, the model is trained sequentially through examples that contain 50 packets (corresponding to 0.5s).  Back-propagation through time is used to compute gradients across the memory. Furthermore, the model is trained to remember by supervising it on objects with 0 points, as long as the object was seen in any of the previous packets. In practice, due to GPU memory constraints, we only compute the forward pass in the first 40 packets to warm-up the memory, then forward and backward through time in the last 10 packets. 

\cutsectionup
\section{Experimental Evaluation}
\label{sec:experimental_results}
\cutsectiondown

We evaluate our model on a real world  dataset for 3D object detection. In particular, we compute mean average precision (mAP)  in the default detection setting (using full \ang{360} sweeps) and propose a new metric that takes into account the latency incurred by different input granularities (i.e., per-packet processing vs. sweep building). Our experimental results show that our model far outperforms the baselines in the per-packet setting while remaining competitive with the state-of-the-art in the full sweep setting. Furthermore, our latency evaluation also uncovers a problem with the mAP metric in the default setting as it does not accurately measure real world performance.

\paragraph{Dataset:}
Since there is no public available dataset that provides packets, we collect a new dataset, \ourdataset, containing  6500 snippets with diverse conditions (e.g., geographical, lighting, road topology, vehicle types). The LiDAR  rotates at a rate of 10hz and emits new packets at 100Hz -- each roughly covering a \ang{36} region -- for a total of 16,250,000 packets  (1,625,000 frames). Accurate ego-pose is  available for each LiDAR packet via a commercial localization system. Labels provide both the spatial extents and motion  of vehicles, cyclists and pedestrians, from which we can extract accurate bounding boxes at  discrete observation times as well as  in continuous time through the use of a precise motion model. Note that if the observation of an instance is split across packets, each packet will have an instance of the label according to the pose  at the timestamp of the packet. 

\begin{figure}[t]
  \centering
  \includegraphics[width=0.7\linewidth]{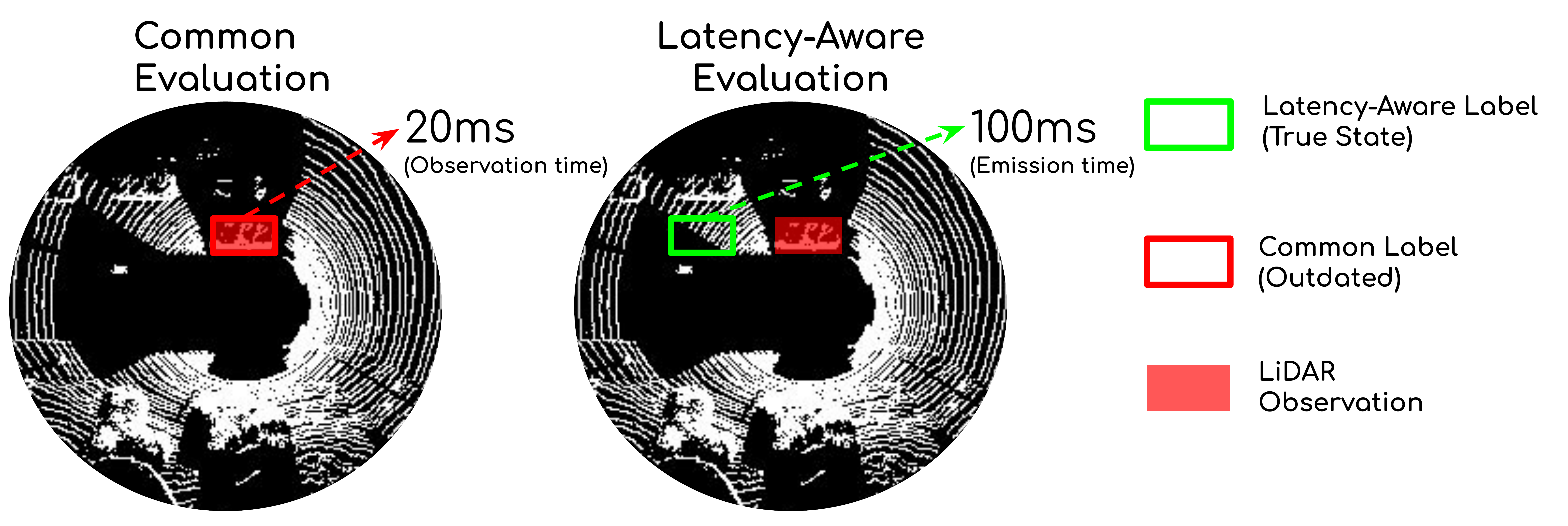}
  \caption{In contrast to the commonly used  mAP, our proposed metric takes into account the latency between observation time and detection emission time.} 
  \label{fig:latency_map}
\end{figure}

\newcolumntype{s}{>{\centering\arraybackslash}X}
\begin{table}[tb]
    \centering
    \begin{tabularx}{0.85\textwidth}{l|r|r|r}
        \toprule
         & Accumulation (ms)& Inference (ms) & Total Latency (ms)\\
        \midrule
                PointRCNN \cite{chen2019fast}                          & 100         & 390         & 490          \\
        PointPillars \cite{lang2019pointpillars,yan2018second} & 100         & 37          & 137         \\
        HDNET \cite{yang2018hdnet}                             & 100         & 28          & 128         \\
        Our \textsc{\ourmodelshort{}}                          & \textbf{10} & \textbf{11} & \textbf{21} \\
        \bottomrule
    \end{tabularx}
    \caption{\textbf{End-to-end Latency:} We report the end-to-end latency in ms of the models as defined by the time it takes to accumulate the data, run inference and decode detections. Accumulation considers a LiDAR operating at 10hz and inference is done with a NVIDIA 2080Ti.}
    \label{table:model_runtimes}
\end{table}

\newcolumntype{s}{>{\centering\arraybackslash}X}
\begin{table}[t]
    \centering
    \begin{tabularx}{\textwidth}{l|ss|ss|ss||ss|ss|ss}
        \toprule
        \multicolumn{1}{s|}{\multirow{3}{*}{Model}} &
        \multicolumn{6}{c||}{\textbf{Packet Stream}} &
        \multicolumn{6}{c}{\textbf{Full Sweep}} \\
        \multicolumn{1}{s|}{} &
          \multicolumn{2}{s|}{Vehicle} &
          \multicolumn{2}{s|}{Pedestrian} &
          \multicolumn{2}{s||}{Cyclist} &
          \multicolumn{2}{s|}{Vehicle} &
          \multicolumn{2}{s|}{Pedestrian} &
          \multicolumn{2}{s}{Cyclist} \\

        \multicolumn{1}{l|}{}                    &       0.5 &      0.7  &       .5m &       .3m &      0.3  &      0.5 &       0.5 &       0.7 &      .5m  &       .3m &       0.3 &       0.5\\\midrule
        HDNET \cite{yang2018hdnet}               &      75.6 &      63.6 &      71.0 &      63.9 &      21.3 &      15.3&      79.6 &      57.8 & {\bf80.2} & {\bf69.8} &      54.6 &      33.8\\
        PointPillars \cite{lang2019pointpillars,yan2018second} &      66.8 &      47.7 &      53.4 &      49.2 &      16.8 &       6.1&      84.2 &      61.1 &      74.4 &      68.9 &      56.1 &      34.9\\
        PointRCNN \cite{chen2019fast}            &      70.2 &      63.1 &      49.3 &      47.5 &      28.4 &      25.9&      72.4 &      57.4 &      54.8 &      52.4 &      31.9 &      26.7\\
        Our \textsc{\ourmodelshort{}}            & {\bf91.8} & {\bf80.5} & {\bf80.3} & {\bf72.5} & {\bf60.8} & {\bf40.7}& {\bf86.4} & {\bf66.4} &      76.7 &      67.8 & {\bf61.0} & {\bf39.5}\\
        
        \bottomrule
    \end{tabularx}
    \caption{\textbf{Latency mAP}: Labels are considered at detection emission times.}
    \label{table:latency_results}
\end{table}

\paragraph{Baselines:} We provide a wide range of baselines that exploit different representations. 
\textbf{HDNET}~\cite{yang2018hdnet} is a detection model that processes input point clouds into occupancy voxels and performs 2D convolution in BEV using the $z$ axis voxels and HD maps as  feature channels.
\textbf{PointRCNN}~\cite{pointrcnn} processes raw LiDAR inputs using a PointNet~\cite{qi2017pointnet} backbone to perform foreground segmentation and generate region-of-interest (RoI) proposals. The RoI proposals are then processed by a classification and bounding box refinement network to output 3D detections.
\textbf{PointPillars}~\cite{lang2019pointpillars,yan2018second} groups input points into discrete bins in BEV and uses PointNet~\cite{qi2017pointnet} to extract features of each bin. The BEV features are then processed with 2D convolutions to generate detection outputs. Note that the PointRCNN and PointPillar baselines do not make use of HD maps.

\paragraph{Metrics:}
We evaluate our method using mean average precision (\emph{mAP}) as our metric with  IOU thresholds of [0.5, 0.7] for vehicles, [0.3, 0.5] for cyclists. For pedestrians, we use the $\ell_2$ distance to centroid with thresholds [0.5m, 0.3m] since we treat the detections as circles with a fixed radius, thus only the centroid is predicted. We evaluate with latency-aware labels that take into account the delay introduced by aggregating consecutive packets (\emph{Latency mAP}). We refer the reader to  \autoref{fig:latency_map} for an illustration of this metric. We re-define the detection label for each object in the scene as its state at detection time (green box), rather than observation time (red box), which does not accurately reflect the current state of the world. The benefits of this metric are twofold: (1) It evaluates how well the detector models the state of the real world and the quality of the information that would be used  by downstream  motion planning, and (2) it allows for a direct comparison with standard detection metrics, thus making apparent the effects of latency. 

\newcolumntype{s}{>{\centering\arraybackslash}X}
\begin{table}[t]
    \centering
    \begin{tabularx}{\textwidth}{l|ss|ss|ss||ss|ss|ss}
        \toprule
        \multicolumn{1}{s|}{\multirow{3}{*}{Model}} &
        \multicolumn{6}{c||}{\textbf{Packet Stream}} &
        \multicolumn{6}{c}{\textbf{Full Sweep}} \\
        \multicolumn{1}{s|}{} &
          \multicolumn{2}{s|}{Vehicle} &
          \multicolumn{2}{s|}{Pedestrian} &
          \multicolumn{2}{s||}{Cyclist} &
          \multicolumn{2}{s|}{Vehicle} &
          \multicolumn{2}{s|}{Pedestrian} &
          \multicolumn{2}{s}{Cyclist} \\

        \multicolumn{1}{l|}{}                    &      0.5  &      0.7  &      .5m  &       .3m &      0.3  &      0.5 &       0.5 &      0.7  &      .5m  &       .3m &      0.3  &      0.5 \\\midrule
        HDNET \cite{yang2018hdnet}               &      75.7 &      63.7 &      71.1 &      64.1 &      21.3 &      15.3& {\bf89.5} & {\bf77.2} & {\bf84.3} & {\bf74.7} & {\bf68.3} & {\bf45.5}\\
        PointPillars \cite{lang2019pointpillars,yan2018second} &      66.9 &      48.0 &      53.5 &      49.3 &      16.9 &       6.2&      84.8 &      70.6 &      74.2 &      69.2 &      56.1 &      36.3\\
        PointRCNN \cite{chen2019fast}            &       70.3 &  63.3 & 49.3        &  47.5         &     28.4      &     25.8            &      73.1 &      66.9 &      54.6 &    	 52.7 &      31.4 &      26.9\\
        Our \textsc{\ourmodelshort{}}            & {\bf91.8} & {\bf80.5} & {\bf80.3} & {\bf72.5} & {\bf60.8} & {\bf40.7}&      87.4 &      76.1 &      76.9 &      69.0 &      61.3 &      41.4\\
        
        \bottomrule
    \end{tabularx}
    \caption{\textbf{Common mAP}: Labels are considered at their observation times.}
    \label{table:common_results}
\end{table}

\paragraph{End-to-end Latency:} 
Since implementations might differ, we did not consider model inference times in the latency aware detection metric. However, it is an important factor in the end-to-end latency for safety since it indicates the minimal amount of time the system would require to be able to recognize an actor, i.e., the time taken for sensor data acquisition, model inference, and emission of a corresponding detection for the actor to donwstream systems. We report end-to-end latency timings in \autoref{table:model_runtimes}; our approach leads to a much faster (on average 6x!) detection emission time. 

\paragraph{Latency-aware Detection:}
 \autoref{table:latency_results} shows our results for \ourdataset{}. In the leftmost setting -- \emph{Packet Stream} -- all models are first trained on detection using LiDAR packets (as opposed to full sweeps) and evaluated using the state of the labels at the time of detection (i.e., green box in \autoref{fig:latency_map}). Our  model far outperforms the baselines, which do not have memory and struggle with partial observations (i.e., a single packet as opposed to the full sweep). In the right portion of the table -- \emph{Full Sweep} -- the models are trained using a traditional full sweep setting and evaluation is done using the label states at the end of the sweep (therefore in the worst case an object could move for 100ms before evaluation).

\paragraph{Latency-unaware Detection:}
We additionally evaluate in the standard object detection setting, not taking into account the sweep building latency and using the labels for each object in the scene at the time of observation (i.e., when the LiDAR  hit the object).  The leftmost columns of \autoref{table:common_results} show the results of the models trained in a packet setting. A key takeaway from these results stems from comparing the numbers in the "Packet Stream" setting between Tables \ref{table:latency_results} and \ref{table:common_results}, which shows that the 10ms latency introduced by accumulating a single packet is negligible in the mAP settings we evaluate, since performance remains  the same. Conversely, comparing the "Full Sweep" setting in Tables \ref{table:latency_results} and \ref{table:common_results} shows considerable degradation in metrics. This indicates that the performance of  full sweep models in the real world would be considerably lower.

\newcolumntype{s}{>{\centering\arraybackslash}X}
\begin{table}[tb]
    \centering
    \begin{tabularx}{0.8\textwidth}{l|ss@{\hskip 1cm}ss@{\hskip 1cm}ss}
        \toprule
        \multicolumn{1}{s|}{\multirow{2}{*}{}} &
          \multicolumn{2}{c}{\hskip -1cm Vehicle} &
          \multicolumn{2}{c}{\hskip -1cm Pedestrian} &
          \multicolumn{2}{c}{\hskip -2mm Cyclist}\\
            \multicolumn{1}{l|}{}          & 0.5       & 0.7       & .5m       &  .3m      & 0.3       &       0.5\\\midrule
            No Memory                      & 75.6      & 63.6      & 71.0      & 63.9      & 21.3      &      15.3\\
            Attention                      & 89.3      & 78.2      & 75.9      & 67.9      & 53.5      &      35.3\\
            No Map                         & 90.6      & 79.9      & 79.3      & 71.8      & 59.6      &      40.5\\
            Our \textsc{\ourmodelshort{}}  & {\bf91.8} & {\bf80.5} & {\bf80.3} & {\bf72.5} & {\bf60.8} & {\bf40.7}\\
            \bottomrule
    \end{tabularx}
    \caption{\textbf{Ablation studies:}   Our multi-scale spatial memory is a critical component in our model. Using maps is beneficial but not critical. Labels are  at detection emission time. }
    \label{table:memory_ablation}
\end{table}

\begin{figure}[t]
  \centering
  \begin{tabular} {@{}c@{\hspace{.1em}}c@{\hspace{.2em}}c@{\hspace{.2em}}c@{\hspace{.2em}}c}
      {} & \textbf{Snippet 1} & \textbf{Snippet 2} & \textbf{Snippet 3} & \vspace{.1em} \\
      \rotatebox[origin=c]{90}{\textbf{t=10ms}} &
      \raisebox{-0.5\height}{\includegraphics[width=0.325\linewidth, trim={12cm, 5cm, 3cm, 5cm}, clip]{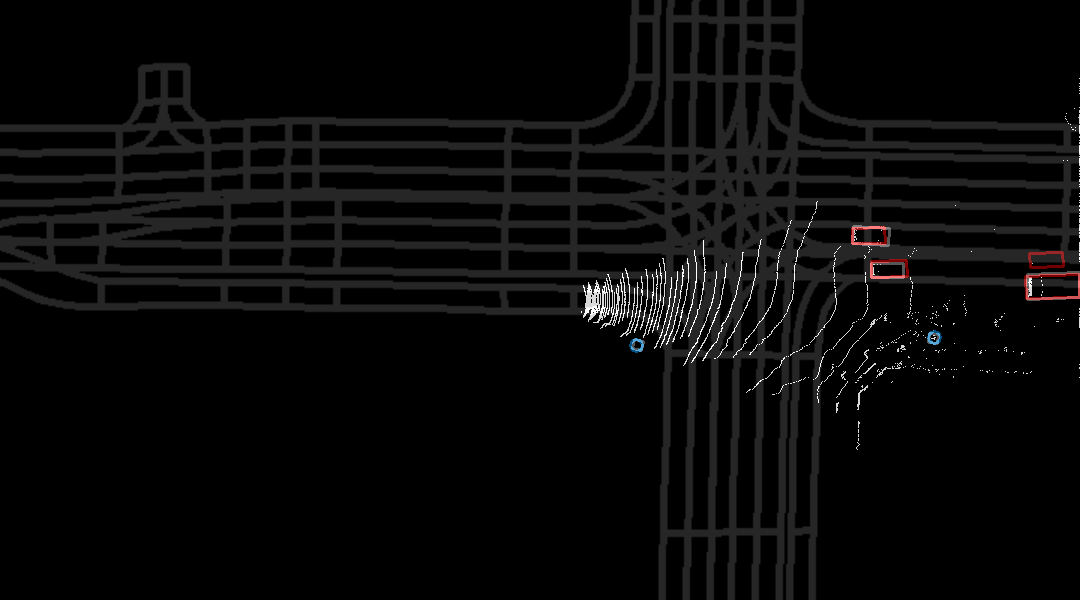}} &
      \raisebox{-0.5\height}{\includegraphics[width=0.325\linewidth, trim={12cm, 5cm, 3cm, 5cm}, clip]{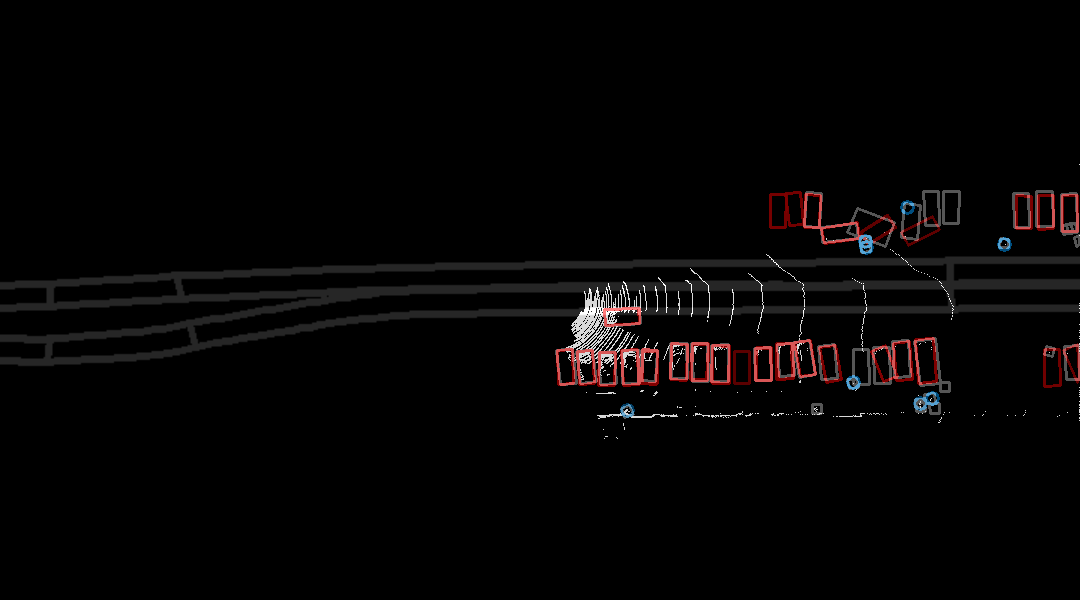}} &
      \raisebox{-0.5\height}{\includegraphics[width=0.325\linewidth, trim={12cm, 5cm, 3cm, 5cm}, clip]{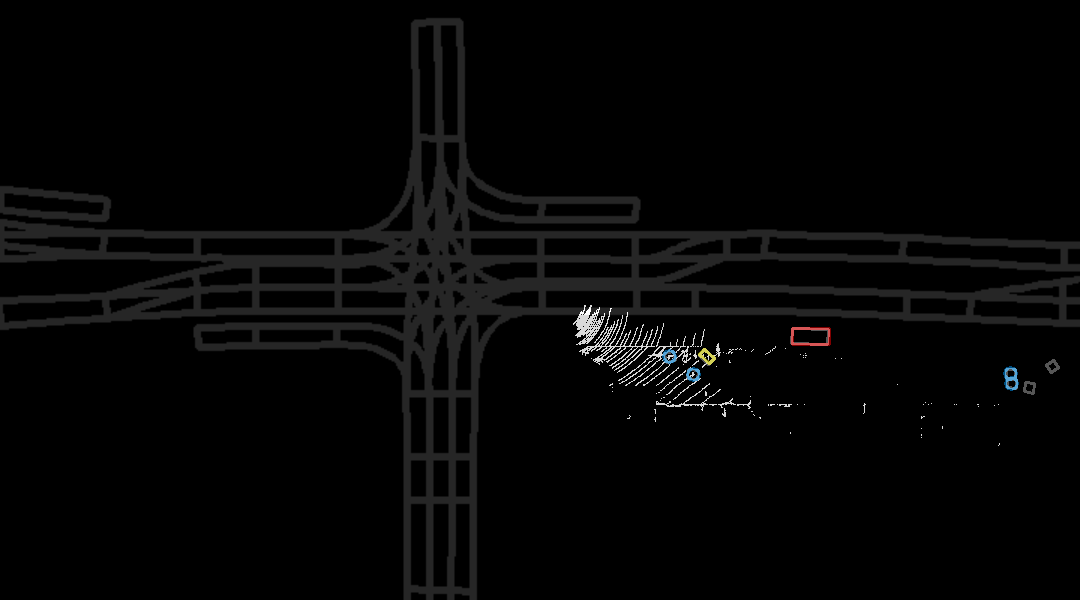}} \vspace{.2em} \\
      \rotatebox[origin=c]{90}{\textbf{t=20ms}} &
      \raisebox{-0.5\height}{\includegraphics[width=0.325\linewidth, trim={12cm, 5cm, 3cm, 5cm}, clip]{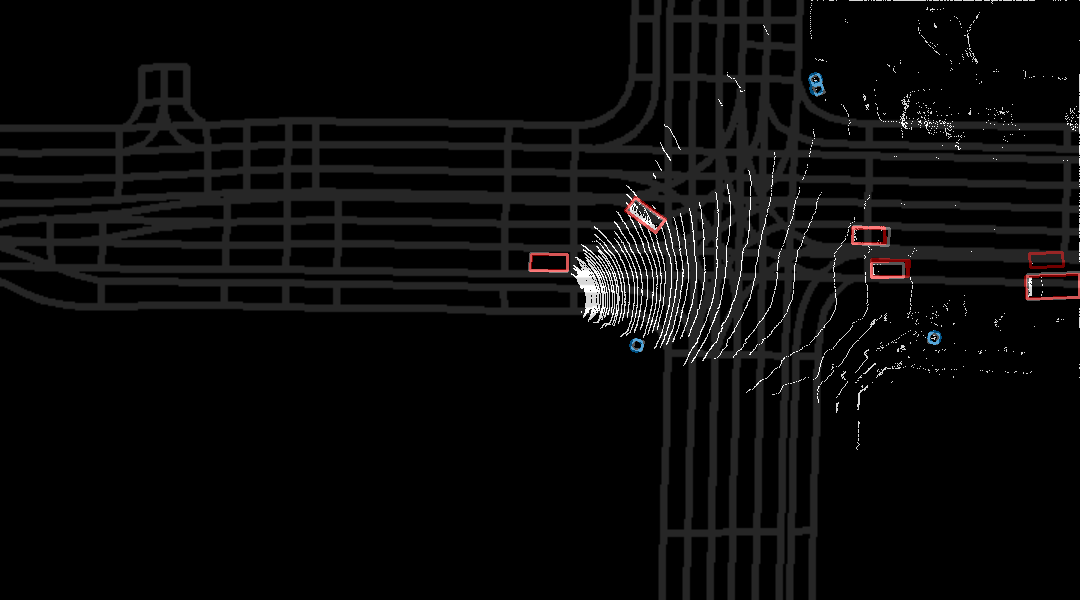}} &
      \raisebox{-0.5\height}{\includegraphics[width=0.325\linewidth, trim={12cm, 5cm, 3cm, 5cm}, clip]{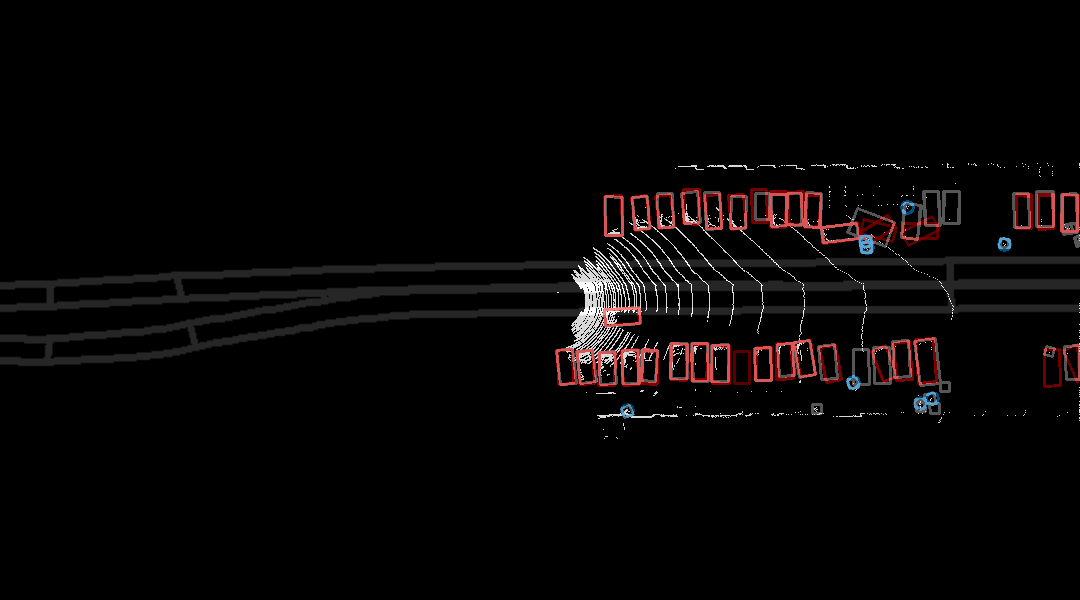}} &
      \raisebox{-0.5\height}{\includegraphics[width=0.325\linewidth, trim={12cm, 5cm, 3cm, 5cm}, clip]{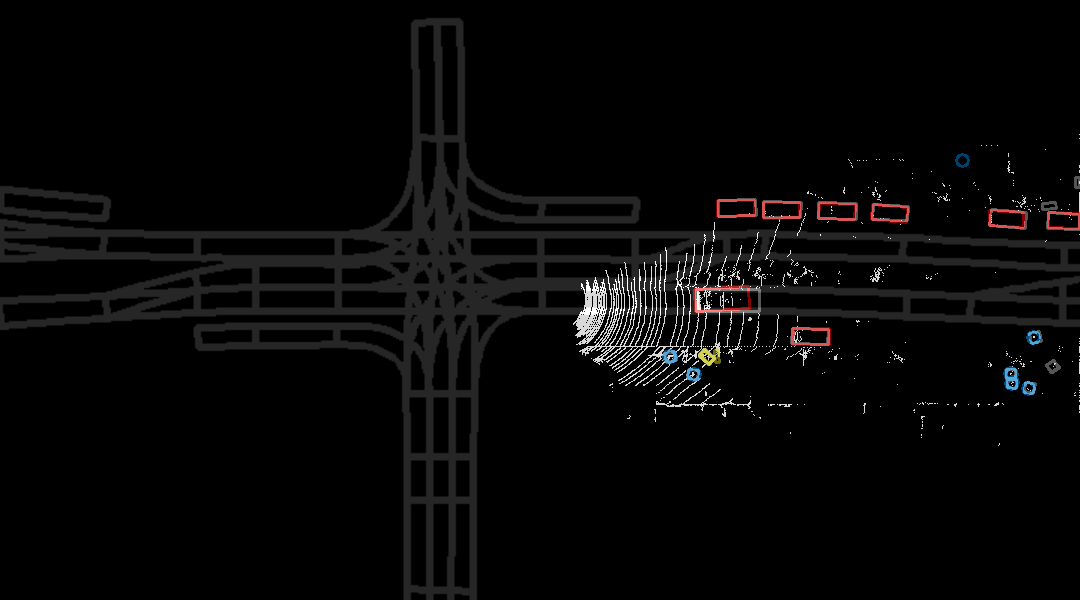}} \vspace{.2em} \\
      \rotatebox[origin=c]{90}{\textbf{t=30ms}} &
      \raisebox{-0.5\height}{\includegraphics[width=0.325\linewidth, trim={12cm, 5cm, 3cm, 5cm}, clip]{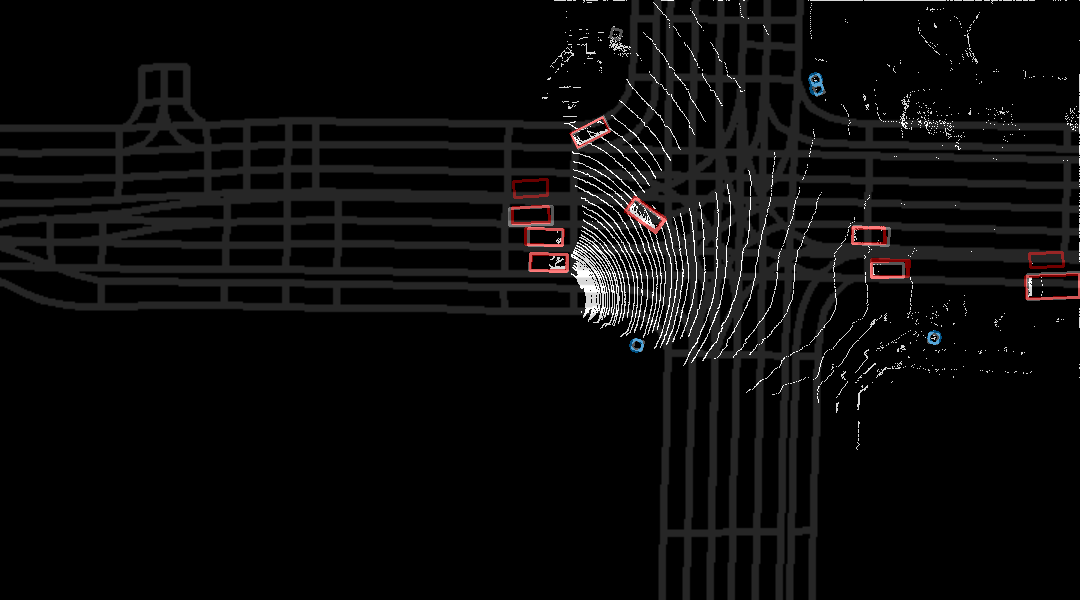}} &
      \raisebox{-0.5\height}{\includegraphics[width=0.325\linewidth, trim={12cm, 5cm, 3cm, 5cm}, clip]{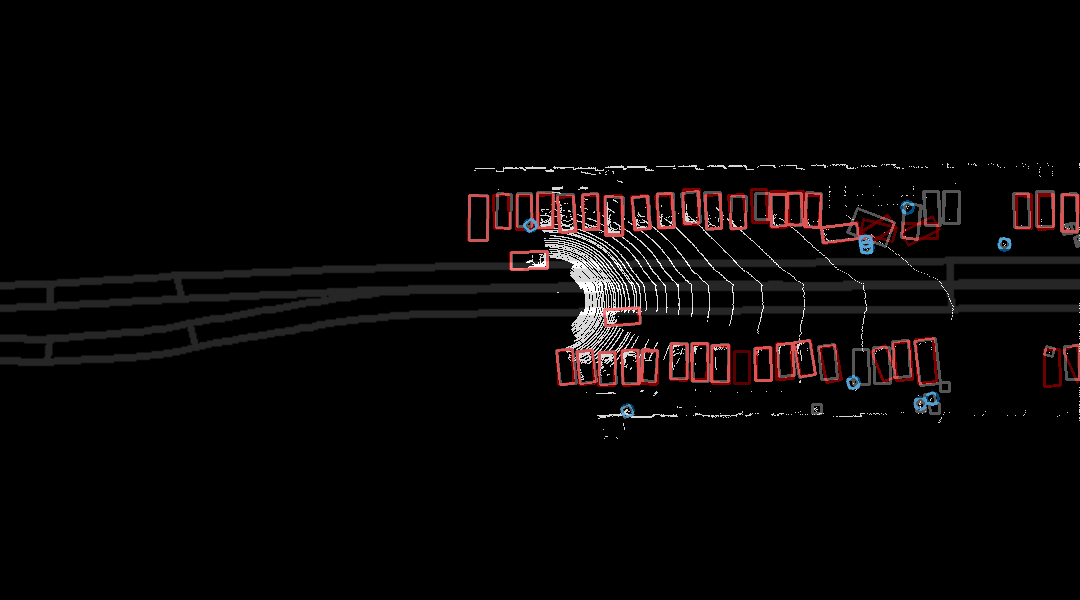}} &
      \raisebox{-0.5\height}{\includegraphics[width=0.325\linewidth, trim={12cm, 5cm, 3cm, 5cm}, clip]{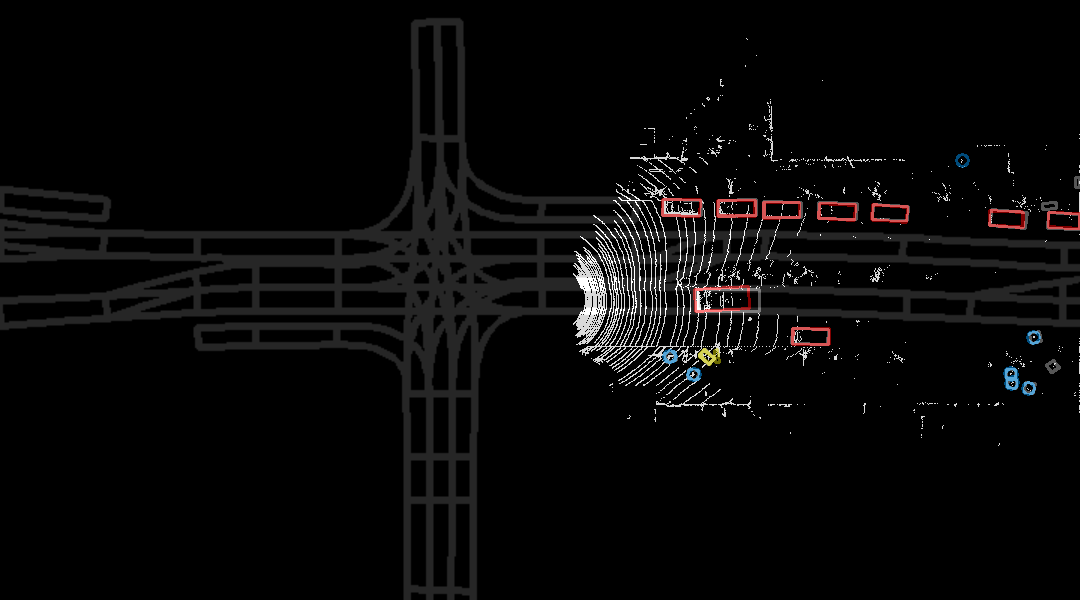}} \vspace{.2em} \\
      \rotatebox[origin=c]{90}{\textbf{t=40ms}} &
      \raisebox{-0.5\height}{\includegraphics[width=0.325\linewidth, trim={12cm, 5cm, 3cm, 5cm}, clip]{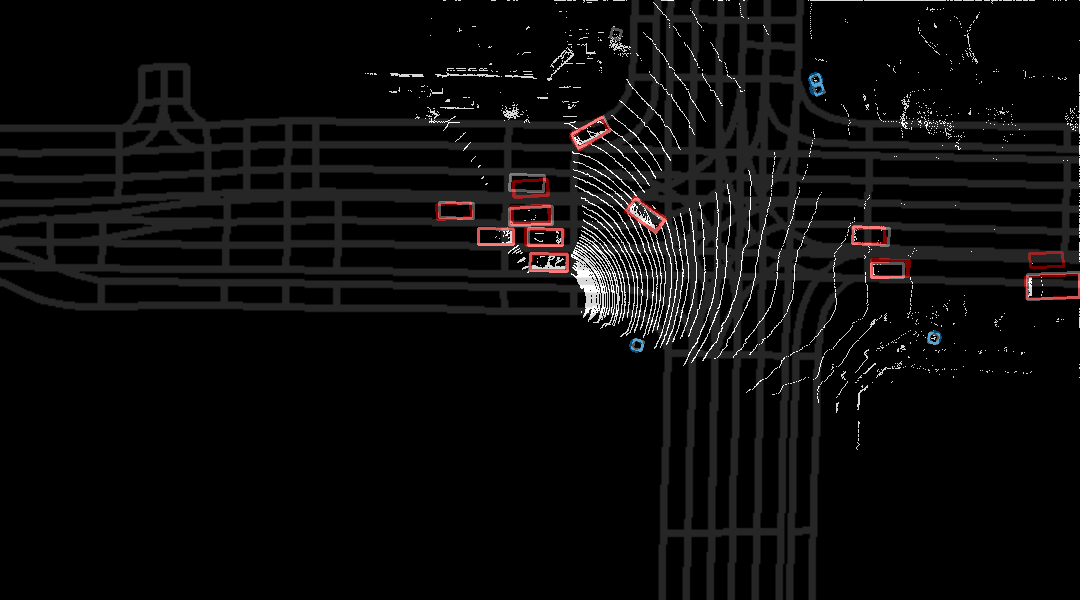}} &
      \raisebox{-0.5\height}{\includegraphics[width=0.325\linewidth, trim={12cm, 5cm, 3cm, 5cm}, clip]{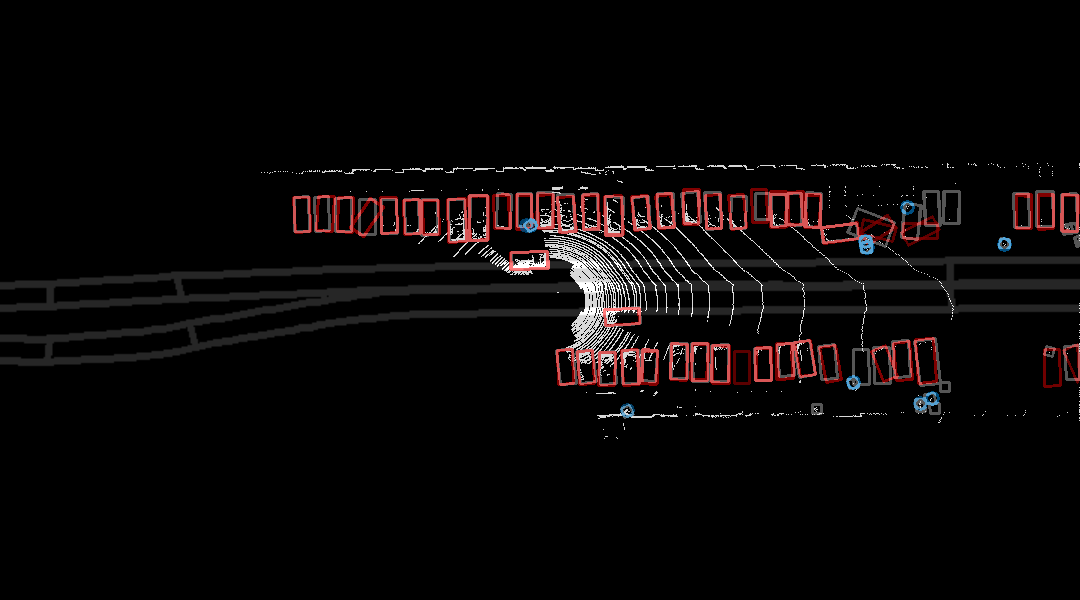}} &
      \raisebox{-0.5\height}{\includegraphics[width=0.325\linewidth, trim={12cm, 5cm, 3cm, 5cm}, clip]{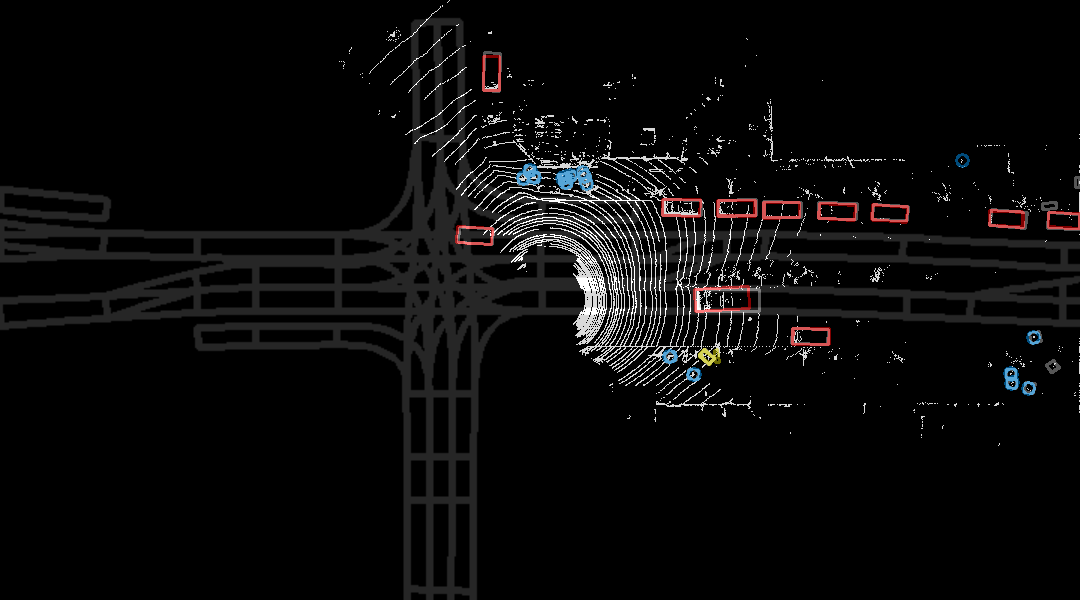}} \vspace{.2em} \\
  \end{tabular}
  \caption{Qualitative results of \textsc{\ourmodelshort{}}. Each column is a sequence of packets from the same snippet. Detected vehicles are shown in red, cyclists in yellow and pedestrians in blue. }
  \label{fig:qualitative}
\end{figure}

\paragraph{Ablation Studies:} We first ablate the memory component of the model. In particular, we evaluate two alternative approaches: (1) \emph{No Memory}: A memoryless instantiation of our model; (2) \emph{Attention:} A memory module that uses linear attention to update the spatial memory (see supplementary for more details). As shown in \autoref{table:memory_ablation}, memory is a fundamental component for effective perception from partial observations. Furthermore, the attention based memory updates are outperformed by our approach which learns the aggregation function through convolutions.
We also evaluate our model without the HD map component to evaluate its importance. The results in \autoref{table:memory_ablation} (\emph{No Map} row) show that while the map backbone proved to be overall beneficial to the model, is not a fundamental component as its removal does not lead to major degradations in metrics.

\paragraph{Qualitative Results:} The qualitative results in \autoref{fig:qualitative} show the predictions of the model over 4 consecutive packets in 3 snippets. The network is able to predict boxes even before points are visible due to the memory module. It can also update the positions of detections as new points arrive to best exploit the evidence. 

\cutsectionup
\section{Conclusion}
\label{sec:conclusion}
\cutsectiondown

We have proposed a novel method for perception of point cloud streaming data. Our approach produces highly accurate object detections at very low latency by using regional convolutions over individual LiDAR packets alongside a spatial memory that keeps track of previous observations. We also introduced a new latency-aware metric that quantifies the cost of data buffering, and how that affects the quality of the models in the real world, which are inevitably affected by latency. Results on the large-scale \ourdataset{} show that our approach far outperforms the state-of-the-art in the packet setting that takes into account latency, while being competitive in the commonly adopted full sweep setting. For future work, we intend to expand the use of the memory module for long term tracking through occlusion and motion forecasting.


\bibliography{references}

\end{document}